%% file: arXiv.tex
\documentclass{article}

\input{Contents/preamble_arxiv}

\begin{document}
\maketitle

\input{Contents/abstract}
\input{Contents/introduction}
\input{Contents/related_works}
\input{Contents/method}
\input{Contents/main_results}
\input{Contents/discussion}
\input{Contents/acknowledgement}

\bibliographystyle{Styles/IEEEbib}
{\small
    \bibliography{References/ref}
}

\end{document}

%% file: Contents/preamble_arxiv.tex
\usepackage{caption}
\usepackage{subcaption}
\usepackage{times}
\usepackage{epsfig}
\usepackage{graphicx}
\usepackage{amsmath}
\usepackage{amssymb}
\usepackage{url}
\usepackage{xcolor}
\usepackage{Styles/spconf}
\title{Exploring the potential of synthetic data to replace real data}
\name{Hyungtae Lee$^{\star,1}$~~~~~~~~~~Yan Zhang$^{\star,2}$~~~~~~~~~~Heesung Kwon$^{3}$~~~~~~~~~~Shuvra S. Bhattacharyya$^{2}$\thanks{© 2024 IEEE. Personal use of this material is permitted. Permission from IEEE must be obtained for all other uses, in any current or future media, including reprinting/republishing this material for advertising or promotional purposes, creating new collective works, for resale or redistribution to servers or lists, or reuse of any copyrighted component of this work in other works.} \vspace{-0.8em}}
\address{$^{\star}$ {\normalsize equal contribution}\vspace{0.5em}\\$^{1}$ BlueHalo~~~~~$^{2}$ University of Maryland, College Park~~~~~$^{3}$ DEVCOM Army Research Laboratory}

%% file: Contents/abstract.tex
\begin{abstract}
\end{abstract}

The potential of synthetic data to replace real data creates a huge demand for synthetic data in data-hungry AI. This potential is even greater when synthetic data is used for training along with a small number of real images from domains other than the test domain. We find that this potential varies depending on (i) the number of cross-domain real images and (ii) the test set on which the trained model is evaluated. We introduce two new metrics, the train2test distance and AP$_\text{t2t}$, to evaluate the ability of a cross-domain training set using synthetic data to represent the characteristics of test instances in relation to training performance. Using these metrics, we delve deeper into the factors that influence the potential of synthetic data and uncover some interesting dynamics about how synthetic data impacts training performance. We hope these discoveries will encourage more widespread use of synthetic data.

\begin{keywords}
Synthetic data, cross-domain, train2test distance, AP$_\text{t2t}$
\end{keywords}

%% file: Contents/introduction.tex
\section{Introduction}
\label{sec:introduction}

Recent advancements in AI have presented surprising performance, often based on very large-scale datasets, \textit{e.g.}, ViT~\cite{ADosovitskiyNeurIPS2021}, LLM~\cite{JDevlinNAACL2019,TBrownNeurIPS2020}. Therefore, much attention is being paid to how to build large-scale datasets. Unfortunately, scaling the dataset to the required extent is challenging, primarily due to the annotation issue. 

One way to meet the need for large-scale datasets is to leverage synthetic data that can be easily scaled with desired annotations. However, using only synthetic data has not yet yielded training performance comparable to using real images due to the domain gap.~\cite{YShenAccess2023,CMaxeyICRA2024,CMaxeyArXiv2024} On the other hand, there are several successful cases of improving training performance using synthetic data with the help of a small number of cross-domain images from a different domain than the test set.~\cite{YShenCVPR2023,HLeeArXiv2024}

\begin{figure}[t]
\centering
\includegraphics[width=.9\linewidth]{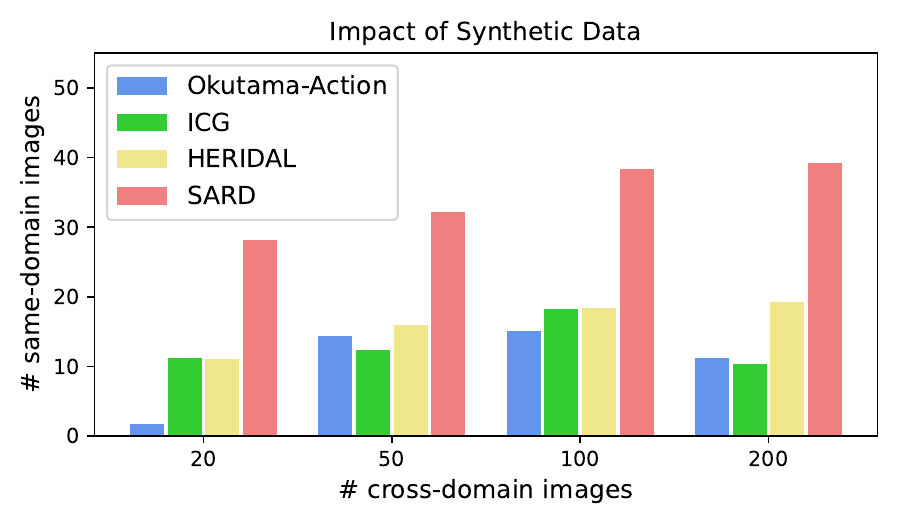}
\vspace{-.5cm}
\caption{{\bf The ability of synthetic data to replace real data.} Each bar shows how much more (same-domain) real data can be replaced when synthetic data is used in training while maintaining the same detection performance. `\# cross-domain image' indicates the number of cross-domain real images used in training, along with synthetic data. The details of measuring the ability of synthetic data are given in Sec.~\ref{ssec:accuracy_match}. }
\label{fig:impact_of_synth}
\end{figure}

In this paper, we aim to explore the ability of synthetic data to replace real data in the cross-domain setting to open up better ways to use synthetic data. We experimentally compare the difference in the ability of cross-domain settings to replace same-domain training images without synthetic data when the cross-domain cases include and do not include synthetic data (Fig.~\ref{fig:impact_of_synth}). This comparison is performed on four datasets, \textit{i.e.}, Okutama-Action~\cite{MBarekatainCVPRW2017}, ICG~\cite{ICGlink}, HERIDAL~\cite{DBozicStulicIJCV2019}, and SARD~\cite{SSambolekAccess2021}), developed for UAV-view human detection. As shown in Fig.~\ref{fig:impact_of_synth}, the impact of using synthetic data in training under the cross-domain setting varies depending on (i) \emph{the number of cross-domain real images}, and (ii) \emph{the test set on which the trained model is evaluated on}.

To further explore these factors that influence the impact of synthetic data, we seek to measure the ability of the training set to represent human instances in the test set as well as background instances that could be potential false positives. \cite{HLeeArXiv2024} introduces a metric that measures the \emph{train2test} distance from the training set to each test instance in the feature space learned using the training set. In this paper, we also introduce a new metric AP$_\text{t2t}$, which calculates average precision over true and false positives according to their train2test distances. Note that training performance is calculated using the occurrence of true and false positives in the test set. This new metric is useful to analyze the impact of using synthetic data in leading the training set to better represent test instances with respect to training performance.

With these metrics (train2test distance and AP$_\text{t2t}$), we carry out a variety of experiments to explore the two major factors defined above that influence the impact of synthetic data. From the experiments, we can reveal the following findings:
\begin{itemize}
\item The ability of synthetic data to replace real data improves as \emph{more real training images are used}.
\item The use of synthetic data affects \emph{medium-confidence detections} greater than other detections..
\item The impact of using synthetic data varies depending on \emph{the test set}, primarily because of its different impact on the occurrence of \emph{false positives} in the test set.
\end{itemize}
These findings can serve as experimental hints to better understand how synthetic data affects training performance.

%% file: Contents/related_works.tex
\section{Related Works}
\label{sec:related_works}

\noindent{\bf Exploring the dynamics of synthetic data.} Recently, there are many studies using synthetic data for training, which also provide simple ablation studies on how to use synthetic data. Here, among these many studies, we will only cover those that provided systematic approaches to explore the dynamics of synthetic data or discovered new aspects that differentiate them from existing studies.

First, there are several approaches, \textit{e.g.}, IS (Inception Score)~\cite{TSalimansNeurIPS2016} and FID (Fr{\' e}chet Inception Distance)~\cite{MHeaselNeurIPS2017}, to compare the diversity and fidelity of synthetic data with real reference data in terms of model representation, \textit{i.e.}, inception v3~\cite{CSzegedyCVPR2015}. Second, many studies~\cite{LFanCVPR2024,YShenArXiv2024,HLeeArXiv2024} explore the scaling behavior of synthetic data regarding its impact in training. Third, there are also several studies that cover new aspects that were not previously thought of. WinSyn~\cite{TKellyCVPR2024} performs three studies on the window element segmentation testbed: i) comparison of the scaling behaviors of real and synthetic data, ii) impact of resolution of synthetic data by stretching one side of the image up to 6K, and iii) impact of category diversity of synthetic data. Lee et al.~\cite{HLeeArXiv2024} and Zhang et al.~\cite{JZhangCVPR2024} explore the domain generalizability achieved by leveraging synthetic data in cross-domain tasks. Shi et al.~\cite{CShiICCV2023} explore the potential of synthetic data to replace real data of visual-language pairs used in CLIP training~\cite{ARadfordICML2021}. Bialer and Haitman~\cite{OBialerCVPR2024} demonstrate that synthetic data (\textit{i.e.}, RadSimReal) can successfully replace real data in applications using radar, which is less complex than the image signal.

Most of the aforementioned studies have focused solely on exploring how specific aspects of synthetic data affect task performance, but lacked analysis of the dynamics of synthetic data behind their effects. We also aim to find the potential of synthetic data to replace real data, similar to Shi et al.~\cite{CShiICCV2023} and Bialer and Hairman~\cite{OBialerCVPR2024}, and at the same time explore the dynamic impact of synthetic data on training through further analysis.\smallskip

\noindent{\bf Bridging synthetic data with real data.} The ability of synthetic data to replace real data depends largely on how the synthetic data is made to appear realistic. One way to acquire realistic-looking synthetic data is to capture properties that exist in the real-world when creating synthetic data. For example, AGORA~\cite{PPatelCVPR2021} and ScoreHMR~\cite{AStathopoulosCVPR2024} obtain synthetic human datasets by fitting human body models into real-world images/videos. Synthetic human datasets can be also made by capturing real physical motions of humans using a motion scanner, \textit{e.g.}, SMPL-X~\cite{GPavlakosCVPR2019}, AMASS~\cite{NMahmoodICCV2019}, BEDLAM~\cite{MBlackCVPR2023}, \textit{etc}. For 1D signals such as radar~\cite{OBialerCVPR2024}, the physical properties of real signals can also be exploited.

Another way to acquire realistic-looking synthetic data is to improve realism in pre-made synthetic data by applying a syn2real transformation, \textit{e.g.}, CycleGAN~\cite{JHoffmanICML2018}, CyCADA~\cite{JHoffmanICML2018}, ObjectFolder 2.0~\cite{RGaoCVPR2022}, or selecting the most realistic synthetic data from the entire synthetic data set, \textit{e.g.}, selection via syn2real feature alignment~\cite{ALiCVPR2024,YYaoCVPR2024}. Our study is performed on synthetic data transformed through a syn2real transformation, but the analysis methods and findings are not limited to this category of synthetic data.\smallskip

\noindent{\bf Training performance evaluation metrics} are essentially defined to measure how well a trained model's decisions or their confidence levels meet the given task's objective. For the discriminative tasks, since AUC~\cite{JHanleyRadiology1982} was first introduced, metrics appropriate for each task have been used in a variety of tasks, \textit{e.g.}, top-$k$ for image classification~\cite{JDengCVPR2009}, AP$^\text{bb}$ (average precision over bounding box) for object detection~\cite{MEveringhamIJCV2010,TLinECCV2014}, mean IOU~\cite{JLongCVPR2015} and AP$^\text{mk}$ (AP over mask)~\cite{KHeTPAMI2020} for segmentation, AP$^\text{triplet}$ (AP over triplets)~\cite{LZhengICCV2015} and CMC (cumulative matching characteristics) top-$k$~\cite{HMoonPerception2001} for re-ID, \textit{etc}. It is noteworthy that in a variety of discriminative tasks, reformulations of AP to suit the tasks, are used. For generative tasks, several metrics have been introduced to measure the extent to which generative output has realistic properties, \textit{e.g.}, IS~\cite{TSalimansNeurIPS2016} and FID~\cite{MHeaselNeurIPS2017}.

In addition to how well it meets the task objectives, several evaluation metrics are introduced to further investigate other aspects of the trained model. For example, FLOPs (Floating-point operations per second) and model size are used for measuring model efficiency and scalability~\cite{HLeeDCS2019}. Robustness in cross-domain tasks are measured with top-1 acc in different domains~\cite{ARadfordICML2021}. For generalizability measure, train2test~\cite{HLeeArXiv2024} and domain gap~\cite{YShenCVPR2023} are introduced. Model efficiency for re-ID can be measured using mINP (mean inverse negative penalty)~\cite{MYeTPAMI2022}. In this paper, we introduce a new metric that falls into the category of measuring additional aspects, AP$_\text{t2t}$, which is designed as a form of AP to analyze how the representation ability of the trained model varies with different confidence levels.

%% file: Contents/method.tex
\section{Methodology}
\label{sec:methodology}

Our goal is to understand the dynamics of how using synthetic data affects training. Given this goal, we do not aim to develop a novel method, but instead to explore what aspects lie behind the impact of synthetic data on model performance. We therefore provide details on (i) how to use synthetic data in training and (ii) how to measure key aspects that influence the impact of synthetic data.\smallskip

\noindent{\bf Leveraging synthetic data in training.} As a method for leveraging synthetic data in training, we use Progressive Transformation Learning (PTL)~\cite{YShenCVPR2023}, which shows a remarkable capability to maximize the impact of synthetic data. In~\cite{YShenCVPR2023}, PTL presents significantly better detection accuracy compared to other methods (\textit{i.e.}, pretrain-finetune, na{\" i}ve merge) that also provide ways for incorporating synthetic data into the training process.

PTL progressively expands a training set by iteratively adding a subset of synthetic data to the set. The subset of synthetic data is constructed in such a way that it includes more synthetic images that are closer to the current training set in terms of the syn2real domain gap. The selected synthetic images are included in the training set after being transformed to look real via the syn2real transformation~\cite{JZhuICCV2017}.\smallskip

\noindent{\bf Measuring the train2test distance.} The expected effect of data augmentation in training is to improve the representation of the trained model on the test set. Accordingly, to explore the extent of the impact of using synthetic data to expand the training data, we need to measure the \emph{train2test} distance between the training set and the test set.

In~\cite{HLeeArXiv2024}, it is derived that the train2test distance can be measured by the Mahalanobis distance, as follows:
\begin{equation}
    d({\bf x}_\text{test}) = \left(f({\bf x}_\text{test})-\mu_\text{train}\right)^\top\Sigma^{-1}_\text{train}\left(f({\bf x}_\text{test})-\mu_\text{train}\right),\label{eq:t2t_distance}
\end{equation}
where $\mu_\text{train}$ and $\Sigma_\text{train}$ are the mean and covariance of the training set in the feature space, respectively, modeled as a multivariate Gaussian distribution. $f(\cdot)$ is the feature encoded by the model trained on the training set. ${\bf x}_\text{test}$ is an instance in the test set. Note that this is only valid if the feature space is formed at the output of the penultimate layer of the model and the model output takes on a sigmoid function\footnote{Most of the advanced models developed for the detection task used in our experiments are sigmoid-based, \textit{i.e.}, RetinaNet~\cite{TLinICCV2017}.}.\smallskip

\noindent{\bf Analyzing true and false positives \textit{w.r.t.} the train2test distance.} To explore the relationship between the representation capability of the training set and training performance, we need to analyze how this capability affects the occurrence of true and false positives in the evaluation. For this analysis, we develop a new metric that measures the change in precision over true positives and false positives based on the train2test distance. Specifically, this metric, called \emph{average precision based on train2test distance} (AP$_\text{t2t}$), is the average precision calculated using the lists of true positives and false positives, respectively, sorted by the train2test distance.\footnote{The conventional AP used as a detection evaluation metric uses lists of true and false positives, respectively, sorted by \emph{detection score}.} A higher AP$_\text{t2t}$ indicates that the training set can represent more true positives in a given test set. 

Formally, let ${\bf X}_\text{tp}$, ${\bf X}_\text{fp}$, and ${\bf X}_\text{fn}$ be sets of true positives, false positives, and false negatives collected from detections\footnote{Detector outputs whose score is greater than the detection threshold are treated as detections. Typically, a very small number, \textit{e.g.} 0.01, is used as the threshold in evaluation, but in our evaluation (section~\ref{ssec:scaling_behavior}), we compare cases with different thresholds, \textit{i.e.}, 0.01, 0.1, or 0.5.}, respectively. Given a score threshold $s_\text{thresh}$, the typical precision $p(\cdot)$ and recall $r(\cdot)$ over detections with scores$\geq s_\text{thresh}$ are defined as:
\begin{eqnarray}
    p(s_\text{thresh}) &=& \frac{|{\bf X}_\text{tp}(s_\text{thresh})|}{|{\bf X}_\text{tp}(s_\text{thresh})|+|{\bf X}_\text{fp}(s_\text{thresh})|},\nonumber\\
    r(s_\text{thresh}) &=& \frac{|{\bf X}_\text{tp}(s_\text{thresh})|}{|{\bf X}|},\label{eq:prec_rec}
\end{eqnarray}
where ${\bf X}_\text{tp}(s_\text{thresh})$ and ${\bf X}_\text{fp}(s_\text{thresh})$ are sets of true and false positives, respectively, with scores greater than or equal to $s_\text{thresh}$, \textit{i.e.}, ${\bf X}_\text{tp}(s_\text{thresh}) = \{{\bf x}\in{\bf X}_\text{tp}~~s.t.~~s_{\bf x}\geq s_\text{thresh}\}$ and ${\bf X}_\text{fp}(s_\text{thresh}) = \{{\bf x}\in{\bf X}_\text{fp}~~s.t.~~s_{\bf x}\geq s_\text{thresh}\}$. {\bf X} is a set of all object-of-interest instances present in the test set, \textit{i.e.}, ${\bf X} = {\bf X}_\text{tp} + {\bf X}_\text{fn}$.

Similarly, precision and recall \emph{based on train2test distance} ($p_\text{t2t}$ and $r_\text{t2t}$) can be calculated by replacing $s_\text{thresh}$ as $d_\text{thresh}$, as follows:
\begin{eqnarray}
    p_\text{t2t}(d_\text{thresh}) &=& \frac{|{\bf X}_{\text{t2t, tp}}(d_\text{thresh})|}{|{\bf X}_{\text{t2t, tp}}(d_\text{thresh})|+|{\bf X}_{\text{t2t, fp}}(d_\text{thresh})|},\nonumber\\
    r_\text{t2t}(d_\text{thresh}) &=& \frac{|{\bf X}_{\text{t2t, tp}}(d_\text{thresh})|}{|{\bf X}|},\label{eq:prec_rec_t2t}
\end{eqnarray}
where ${\bf X}_{\text{t2t, tp}}(d_\text{thresh}) = \{{\bf x}\in{\bf X}_\text{tp}~~s.t.~~d({\bf x})\leq d_\text{thresh}\}$ and ${\bf X}_{\text{t2t, fp}}(d_\text{thresh}) = \{{\bf x}\in{\bf X}_\text{fp}~~s.t.~~d({\bf x})\leq d_\text{thresh}\}$. $d(\cdot)$ can be calculated as in eq.~\ref{eq:t2t_distance}. Note that the inequality changes when constructing ${\bf X}_{\text{t2t, tp}}$ and ${\bf X}_{\text{t2t, fp}}$ because a \emph{smaller} $d$ indicates that the detection can be \emph{better} represented by the training set, whereas a \emph{higher} $s$ represents \emph{higher} confidence for the corresponding detection.

Then, AP$_\text{t2t}$ can be calculated using $p_\text{t2t}$ and $r_\text{t2t}$ as below:
\begin{eqnarray}
    \text{AP}_\text{t2t} &=& \sum_{d\in{\bf D}}{p_\text{t2t}(d) \Delta r_\text{t2t}(d)}\nonumber\\
    &=& \frac{1}{|{\bf X}|}\sum_{d\in{\bf D}}{\frac{|{\bf X}_{\text{t2t, tp}}(d)|}{|{\bf X}_{\text{t2t, tp}}(d)|+|{\bf X}_{\text{t2t, fp}}(d)|}},
\end{eqnarray}
where ${\bf D}$ is the set of train2test distances for all true positive detections, \textit{i.e.}, ${\bf D} = \{d({\bf x}),~~{\bf x}\in{\bf X}_\text{tp}\}$.

%% file: Contents/main_results.tex
\section{Results and Analysis}
\label{sec:results}

\begin{figure*}[t]
\centering
\includegraphics[width=0.235\linewidth]{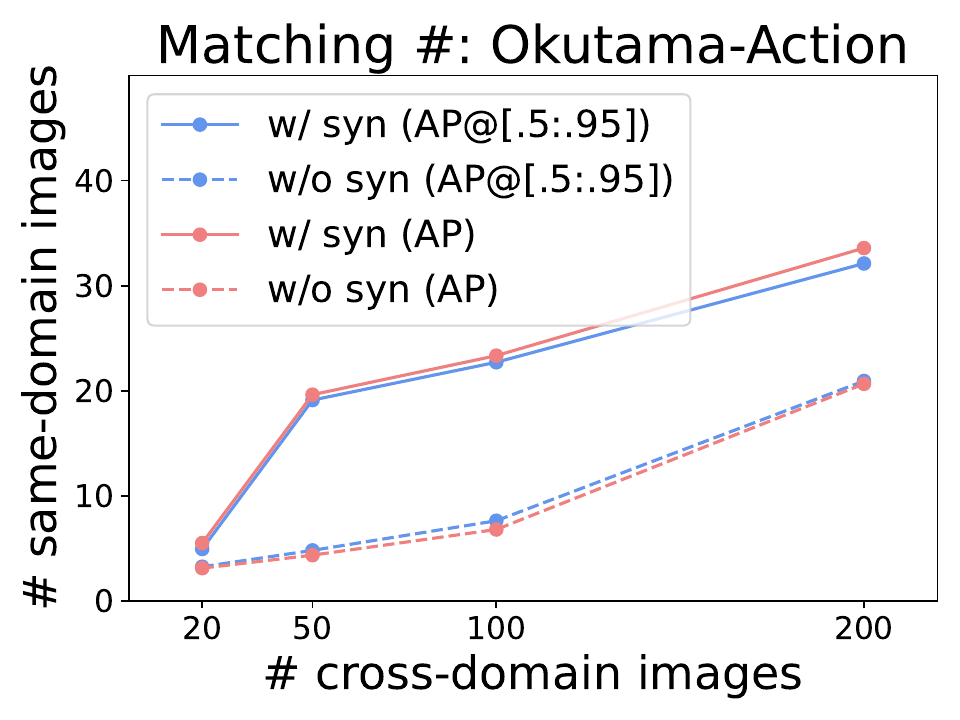}~~
\includegraphics[width=0.235\linewidth]{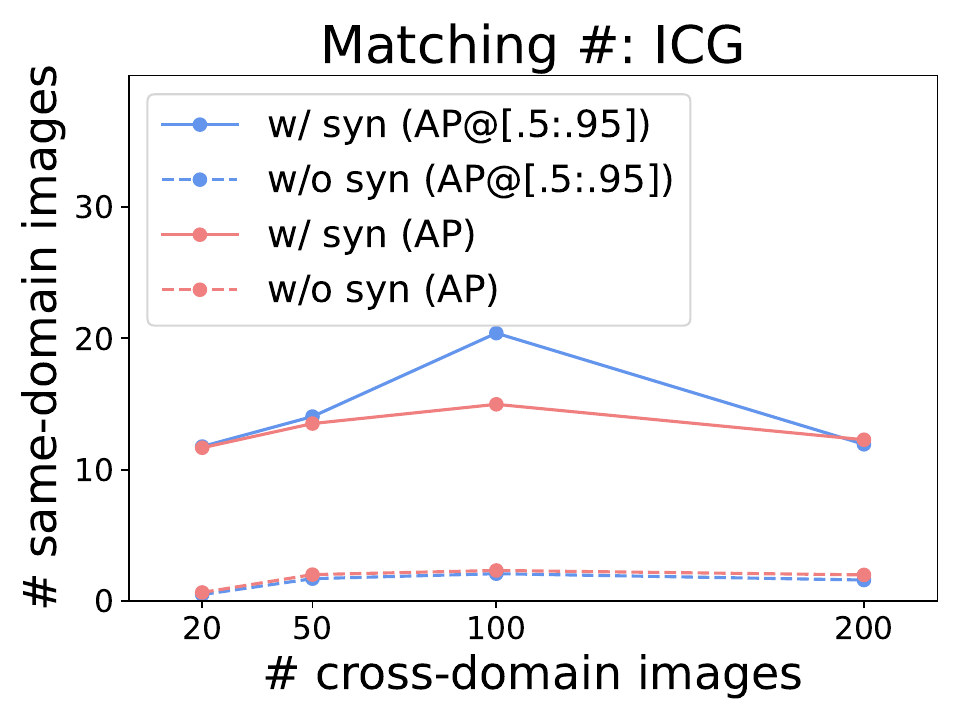}~~
\includegraphics[width=0.235\linewidth]{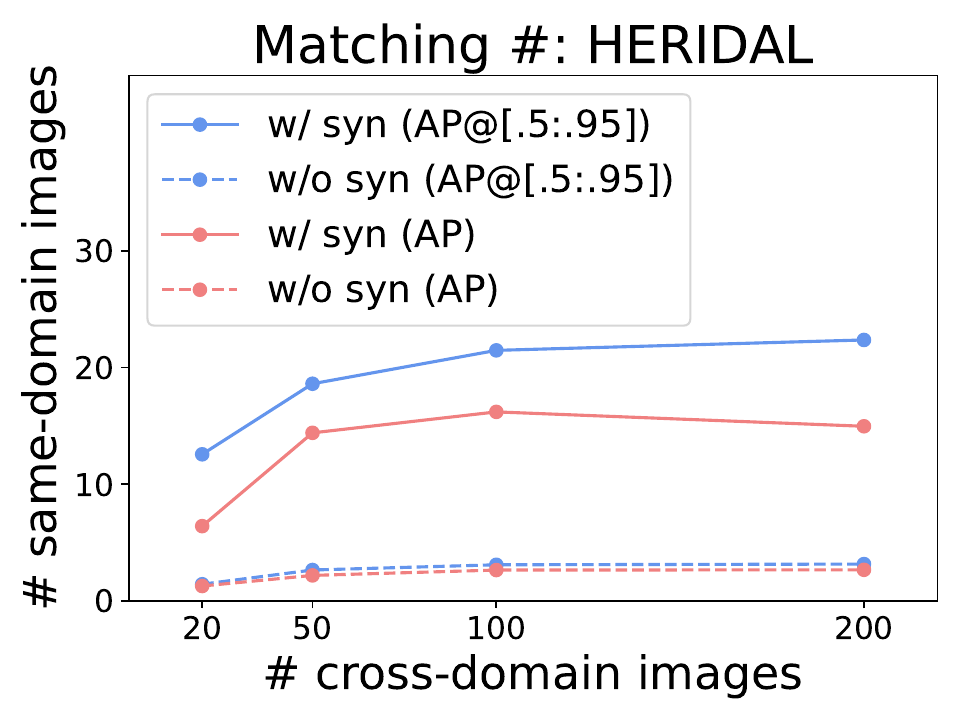}~~
\includegraphics[width=0.235\linewidth]{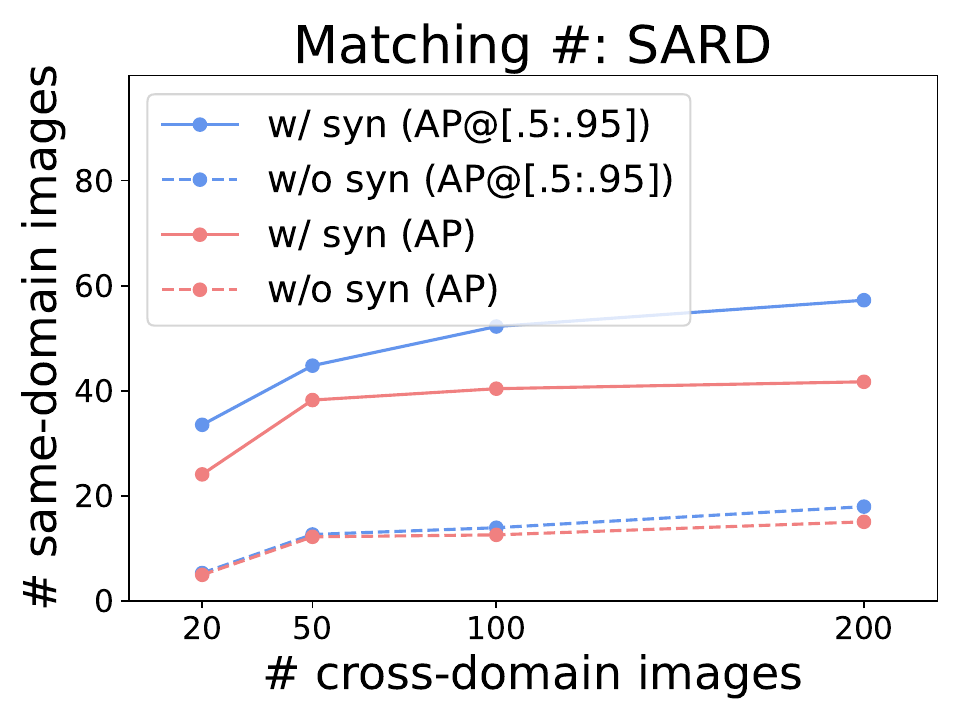}
\vspace{-0.3cm}
\caption{{\bf Number of images from the same- and cross-domain providing equivalent training performance.} The figures show the matching numbers for both with and without synthetic data cases. The increases in matching numbers when using synthetic data are shown in Fig.~\ref{fig:impact_of_synth}. For all experiments, the average of three runs is reported to address random effects that may arise when choosing a specific number of real cross-domain training images.}
\label{fig:num_img_4datasets}
\end{figure*}

\subsection{Setting}
\label{ssec:setting}

\noindent{\bf Task.} The task used in the experiments is UAV-view human detection in which human appearance varies greatly from the UAV perspectives so there is a large need for additional synthetic data. We use the commonly used evaluation metrics AP and AP\@[.5:.95] for the detection task. We also employ the train2test distance and AP$_\text{t2t}$ to measure the representation capability of the training set.\smallskip

\noindent{\bf Dataset.} For real data, we use five datasets constructed for UAV-view human detection: VisDrone~\cite{PZhuTPAMI2021}, Okutama-Action~\cite{MBarekatainCVPRW2017}, ICG~\cite{ICGlink}, HERIDAL~\cite{DBozicStulicIJCV2019}, and SARD~\cite{SSambolekAccess2021}. For synthetic data, we use Archangel-Synthetic~\cite{YShenAccess2023}, which includes humans with diverse poses from various UAV perspectives.

For the cross-domain case, we use VisDrone for training and the other four real datasets for testing. For same-domain experimentation, both training and testing are performed on each of the four datasets other than VisDrone.\smallskip

\noindent{\bf Model training.} For model training, we follow all training specifications used in the original PTL paper~\cite{YShenCVPR2023}. One specification to note is the use of 5 PTL iterations.

\subsection{Match Accuracy between Cross- and Same-domains}
\label{ssec:accuracy_match}

\noindent{\bf Plot design.} We aim to present a comparison based on detection accuracy between the cross-domain and same-domain\footnote{Throughout the experiments, the same-domain setting does not include synthetic data in the training set.} tasks with the scaling up of the training dataset. To achieve this, we generate plots in Figs.~\ref{fig:impact_of_synth} and~\ref{fig:num_img_4datasets} illustrating the number of real images matched to provide comparable detection accuracy across the same- and cross-domains.

We consider four cross-domain settings with different numbers of real images, \textit{i.e.}, 20, 50, 100, and 200. It is very computationally intensive to accurately search for the matching number of same-domain images to each counterpart cross-domain setting. So, we calculated the accuracy of the same-domain settings for every multiple of 5 images and then used interpolation to estimate the matching number.\smallskip

\noindent{\bf Analysis.} From these comparative plots in Figs.~\ref{fig:impact_of_synth} and~\ref{fig:num_img_4datasets}, we discovered two general observations. First, in both scenarios with and without synthetic data, a larger number of cross-domain images can generally replace a larger number of same-domain images. However, when the number of cross-domain images becomes very large (\textit{e.g.}, 200), there is a plateau where this increasing trend slows down or even declines. Second, using synthetic data in training significantly improves performance in all cross-domain cases while the impact varies across different test datasets.

From these observations, we would like to ask the following question: \emph{Why does the impact of synthetic data in training vary depending on (i) the amount of real training data and (ii) the test dataset?} In the next two subsections, we aim to answer each question by exploring, through experimental evidence, the dynamics by which these factors influence the impact of synthetic data.

\begin{figure*}[t]
\centering
\includegraphics[width=0.235\linewidth]{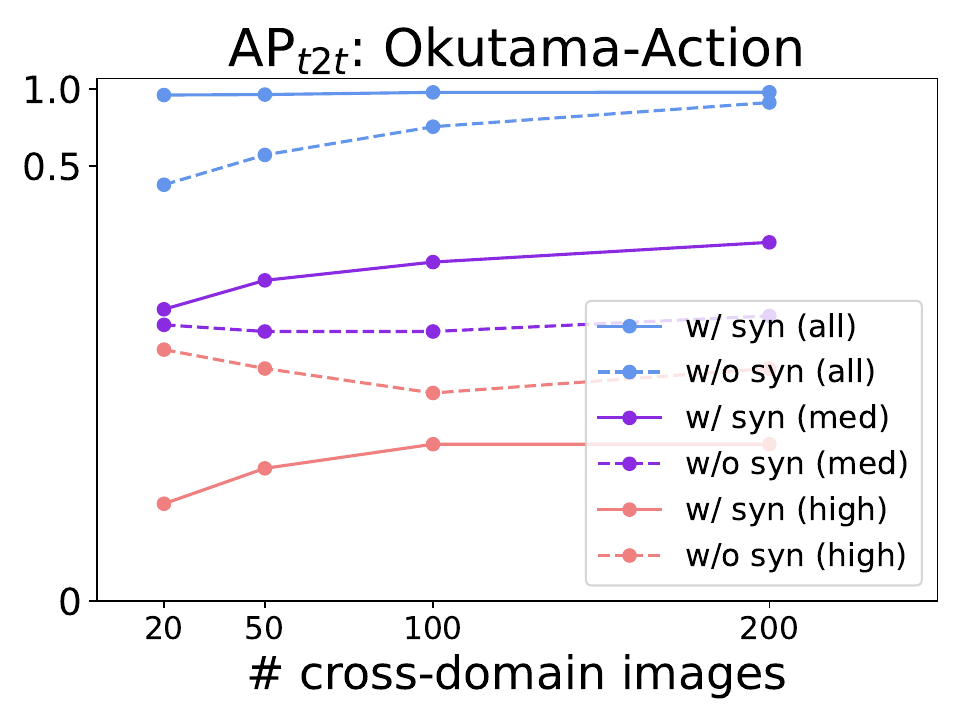}~~
\includegraphics[width=0.235\linewidth]{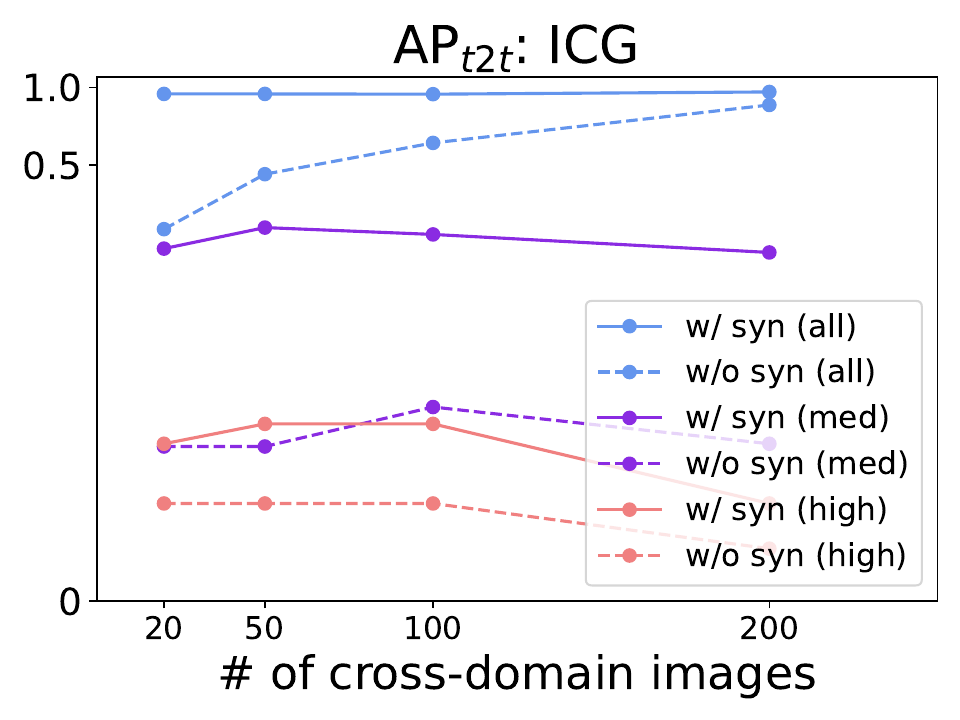}~~
\includegraphics[width=0.235\linewidth]{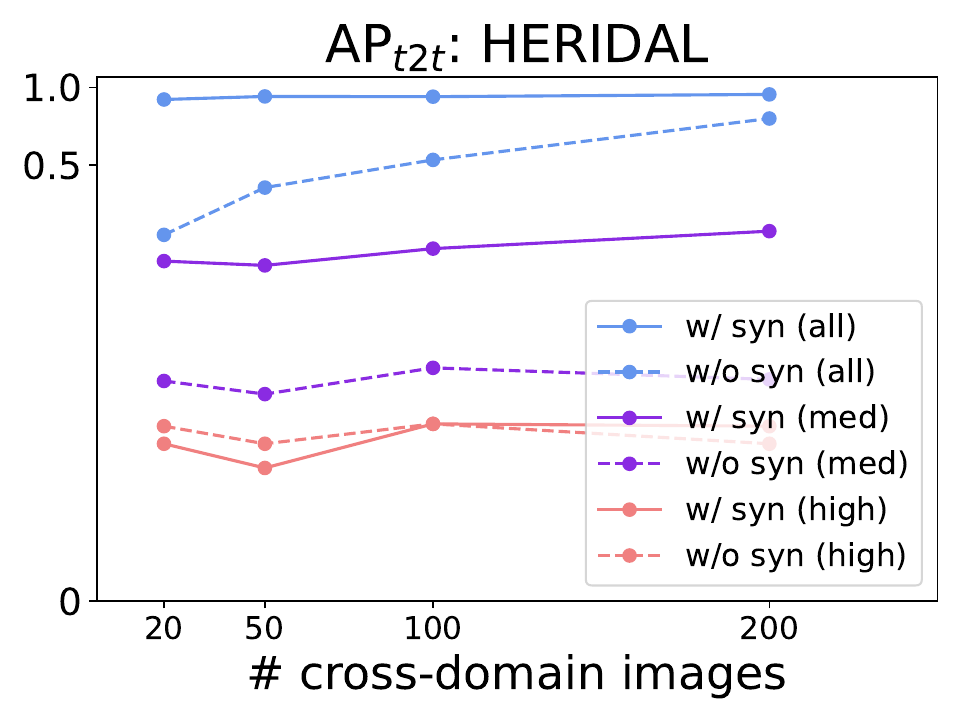}~~
\includegraphics[width=0.235\linewidth]{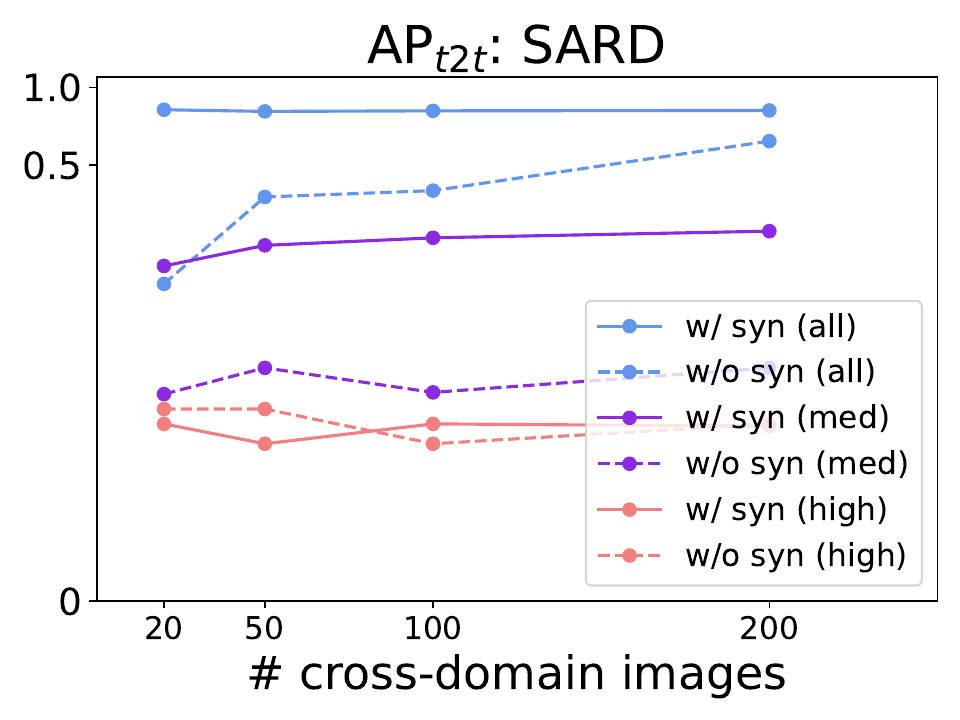}
\vspace{-0.3cm}
\caption{{\bf Scaling behavior of real training data in terms of AP$_\text{t2t}$.} Here, `high', `med', and `all' represent the high-confidence detections, the above-medium-confidence detections, and all potential detections with minimum confidence score, respectively. The $y$-axis is shown in a logarithmic scale to better focus on the scaling behavior seen at low APs.}
\label{fig:scaling_behavior}
\end{figure*}

\subsection{Scaling Behavior of the Cross-domain Training Set}
\label{ssec:scaling_behavior}

\noindent{\bf Plot design.} Our aim is to find out how the impact of synthetic data on enhancing the representation ability of the training set depends on the number of real images in the training set. Accordingly, we design plots that compare the scaling behavior of the real training set in terms of AP$_\text{t2t}$ with and without synthetic data, as shown in Fig.~\ref{fig:scaling_behavior}. 

The AP$_\text{t2t}$ value can vary depending on the detection score threshold used to determine detections from which AP$_\text{t2t}$ is calculated. We consider three sets using the thresholds of 0.01, 0.1, and 0.5, respectively, which include all potential detections, detections with high confidence, and detections with above-medium confidence.\smallskip

\noindent{\bf Analysis.} For the all potential detections, an increasing scaling behavior is observed when synthetic data is not used in training, but the increasing behavior appears to be saturated when using synthetic data. From these observations, more training data may have a greater effect on enhancing the representation ability of the training set, which is not surprising; our experiments here give quantitative insight into this expected trend. In addition, \emph{using synthetic data in training reduces the train2test distance for most positive instances without having the same effect on false instances, resulting in nearly perfect AP$_\text{t2t}$ performance}.

For the above-medium-confidence detections, AP$_\text{t2t}$ increases as more real images are used in training only when synthetic data is also used in training. It is also observed that AP$_\text{t2t}$ consistently increases with the use of synthetic data in all cases. This indicates that \emph{using synthetic data ensures that the training set has the ability to represent more positive (test) cases, and this ability improves as more real images are included in the training set}. Remarkably, for these above-medium-confidence detections, the scaling behavior of the real training data for AP$_\text{t2t}$ is most similar to that for detection accuracy (Fig.~\ref{fig:num_img_4datasets}).

For the high-confidence detections, AP$_\text{t2t}$ is generally low, with no significant difference depending on the number of real images regardless of whether synthetic data is used. This can be interpreted as \emph{many false positives having features very similar to those represented for the object-of-interest by the training set, resulting in a small train2test distance}. In addition, the impact of synthetic data in improving the representation ability of the training set is minimal.

In conclusion, \textbf{the positive effect of synthetic data in improving the representation ability of the training set is evident specifically in the case of medium-confidence detections.}

\begin{figure*}[t]
\centering
\includegraphics[width=0.235\linewidth]{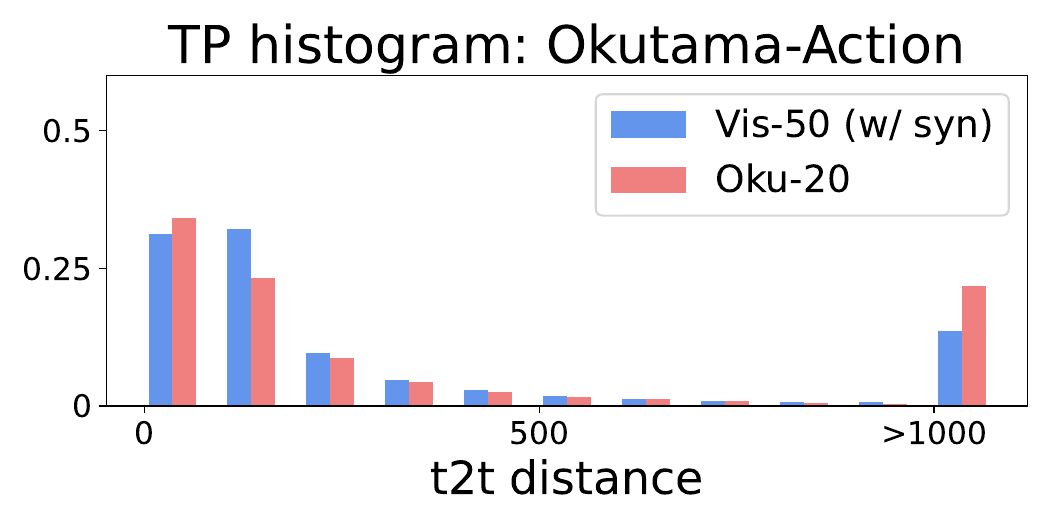}~~
\includegraphics[width=0.235\linewidth]{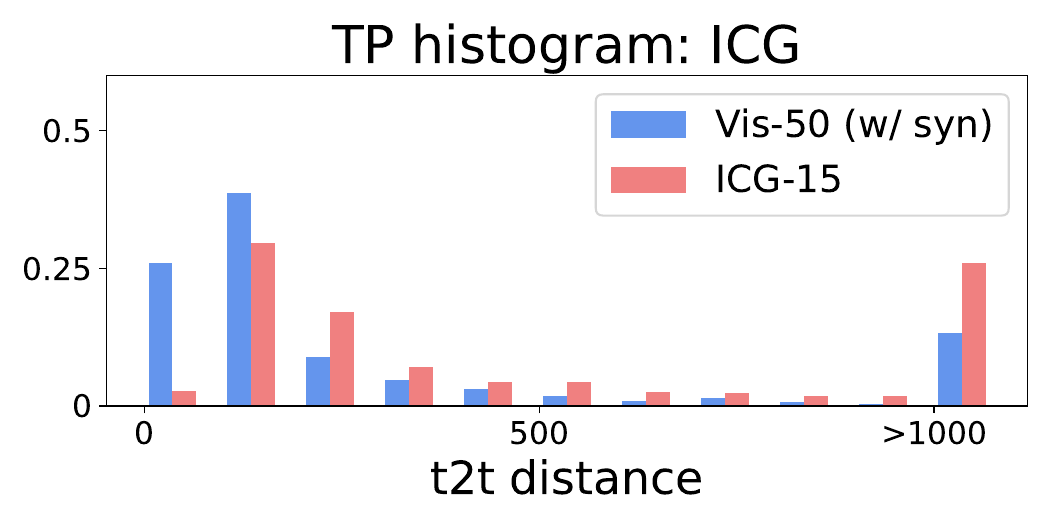}~~
\includegraphics[width=0.235\linewidth]{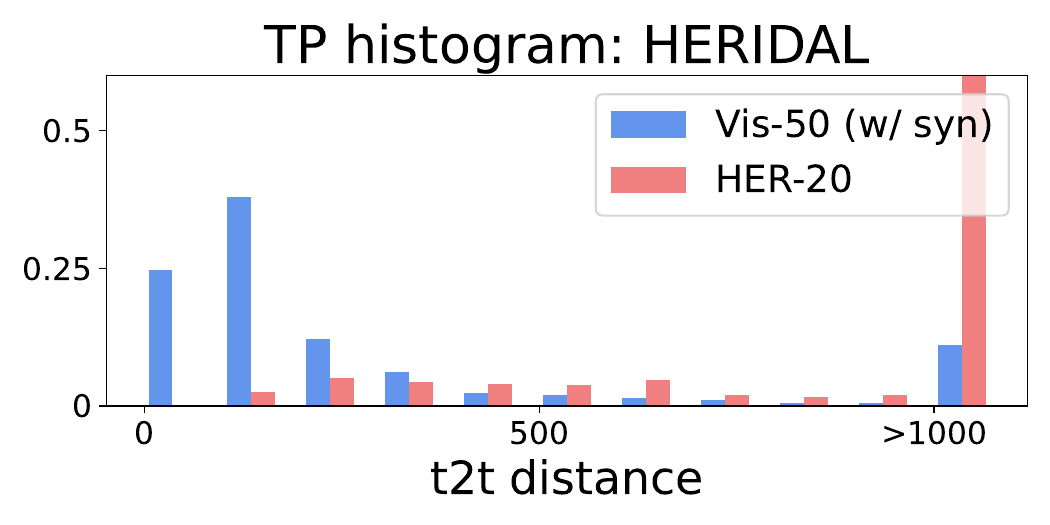}~~
\includegraphics[width=0.235\linewidth]{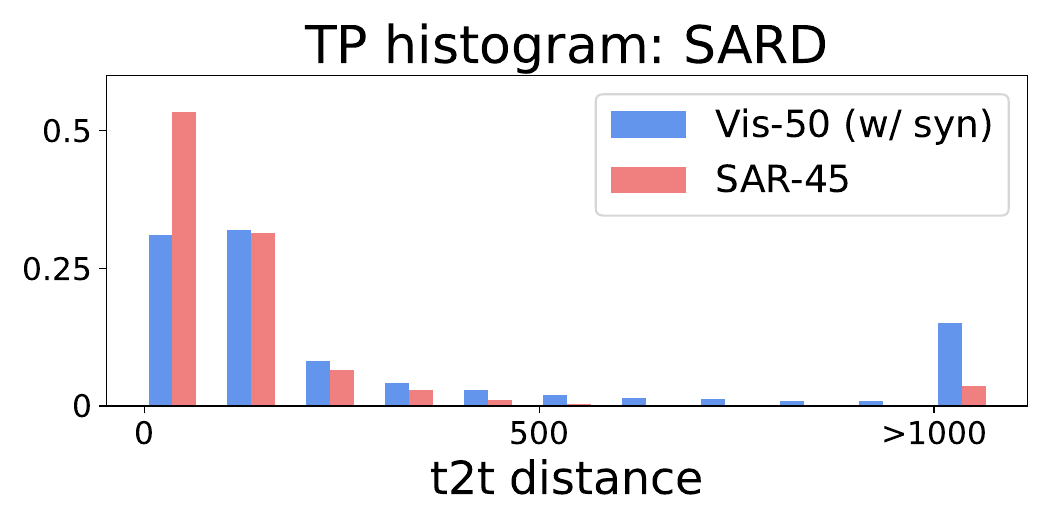}\\
\includegraphics[width=0.235\linewidth]{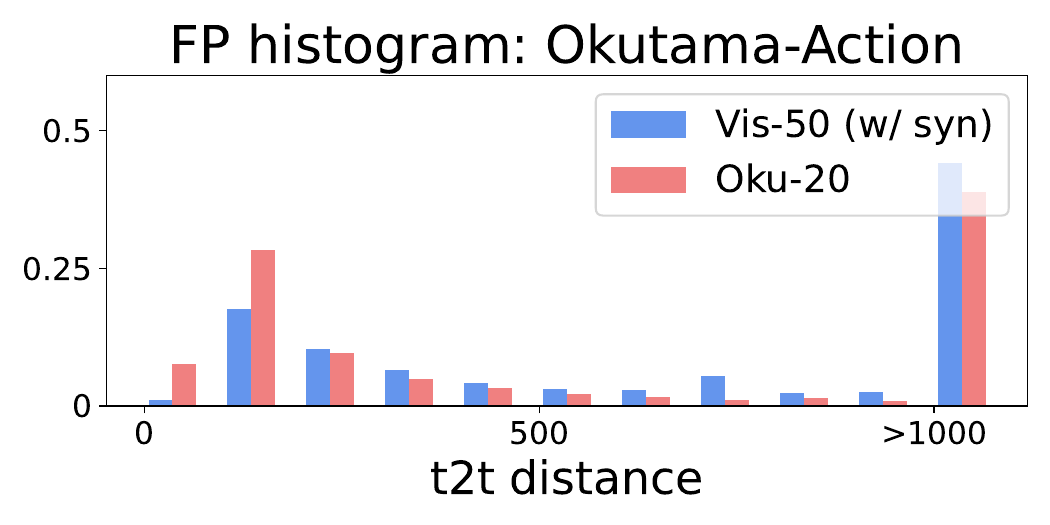}~~
\includegraphics[width=0.235\linewidth]{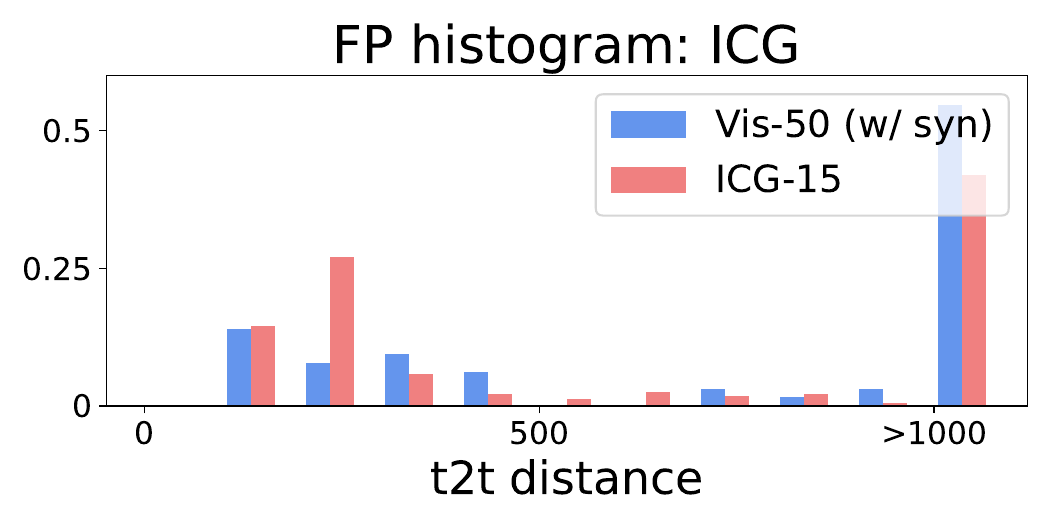}~~
\includegraphics[width=0.235\linewidth]{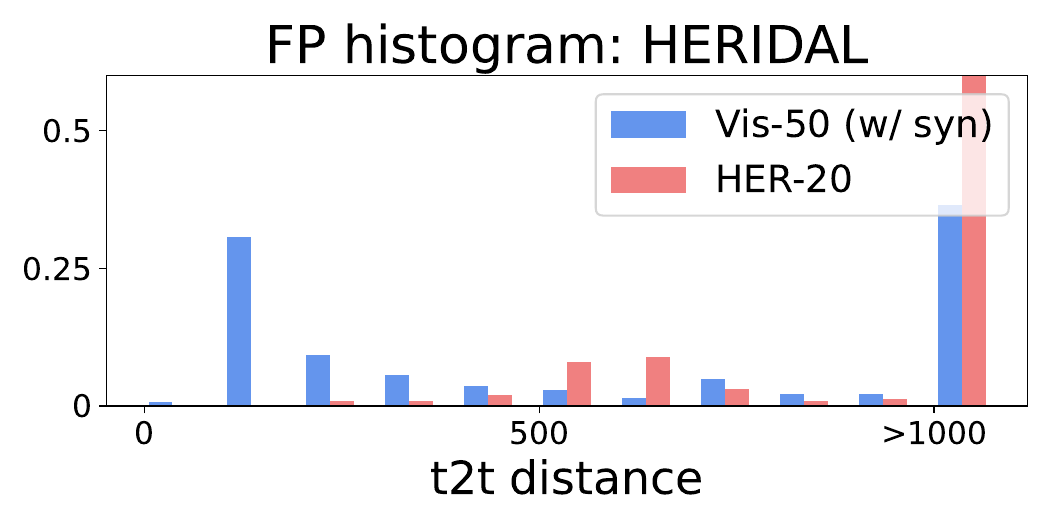}~~
\includegraphics[width=0.235\linewidth]{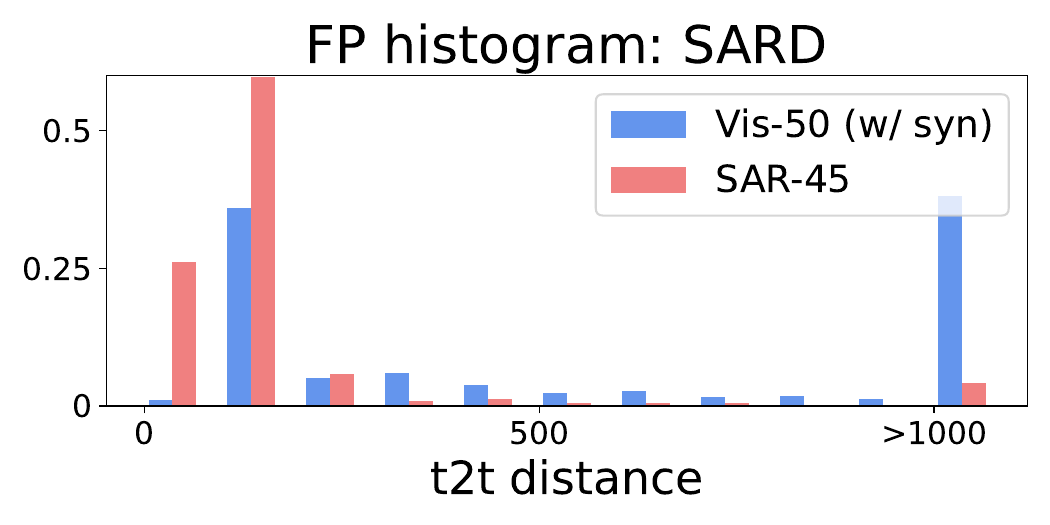}
\vspace{-0.3cm}
\caption{{\bf train2test distribution of TP and FP.} The top and bottom rows show the histograms for TP and FP, respectively. Each setting represents ``XXX-N", where XXX is an abbreviation for the training dataset name and N is the number of real training images, \textit{e.g.}, Vis-50 refers to a training set containing 50 VisDrone images.}
\label{fig:histogram}
\end{figure*}

\subsection{Impact of Synthetic Data Depending on the Test Set}
\label{ssec:testset_effect}

\noindent{\bf Plot design.} We explore the impact of using synthetic data on performance depending on which test set the trained model is evaluated on. For this purpose, we plot the train2test distribution of true positives (TP) and false positives (FP) when using synthetic data, as shown in the top and bottom rows of Fig.~\ref{fig:histogram}, respectively. We also include the distributions in the same-domain case as a comparison group. 
This comparison is performed using a case involving 50 cross-domain real images and a same-domain case that provides matching accuracy.\smallskip

\noindent{\bf Analysis.} A notable observation from this comparison is that when using a cross-domain training set, the train2test distribution for TP is similar regardless of the test set (blue bars in the top rows). On the other hand, the other distributions appear differently for each test set. 

These observations indicate that (i) \emph{the ability of synthetic data to reduce the train2test distances for human instances on the test set is consistent regardless of the test set}, (ii) \emph{the effect of reducing the train2test distances of FP by using the synthetic data depends on the test set}, and (iii) \emph{this ability when using a same-domain training set manifests itself differently depending on which train/test set-pair is used}. 

Therefore, it can be concluded that \textbf{the difference in the ability of synthetic data to replace real data across test sets is because this ability handles false positives differently for each test set}. In addition, the differences in the training ability of each same-domain set, as indicated by different train2test distributions for TP and FP, also affect the matching number between the cross- and same-domain cases, which appears to vary across test sets (our main experiment shown in Figs.~\ref{fig:impact_of_synth} and~\ref{fig:num_img_4datasets}).

%% file: Contents/discussion.tex
\section{Discussion}
\label{sec:discussion}

We summarize the findings from our experiments as follows:
\begin{itemize}
\item \emph{The use of synthetic data has the greatest impact in enhancing a cross-domain training set to be better representative, especially for medium-confidence detections.}
\item \emph{The ability of cross-domain sets to replace real data varies across test sets because the test sets have different potential false positives.}
\end{itemize}
Detection results for true positives fail to achieve high confidence scores when data characteristics are incompletely captured due to insufficient representation ability of the training set. Synthetic data appears to serve to diversify the characteristics of objects-of-interest that can be represented in the training set. In addition, false positives may occur due to insufficient quantity or variety of negative examples in the training set. 

Understanding these dynamics of synthetic data that affect model training can provide important clues for constructing synthetic data while maximizing its ability to take the place of real data. For example, it may be worth considering factors that vary the characteristics of the object-of-interest, such as different human poses, different occluded situations, \textit{etc.}, or adopting diverse backgrounds that provide a variety of negative training examples. We hope that these findings will contribute to more frequent and more informed use of synthetic data in a wider range of research fields.\smallskip

%% file: Contents/acknowledgement.tex
%\section{Acknowledgements}
%\label{sec:acknowledge}

\noindent{\small {\bf Acknowledgements.} This research was sponsored by the Defense Threat Reduction Agency (DTRA) and the DEVCOM Army Research Laboratory (ARL) under a cooperative agreement \\
(W911NF2120076). This research was also sponsored in part by the Army Research Office and Army Research Laboratory (ARL) (W911NF2110258). The views and conclusions contained in this document are those of the authors and should not be interpreted as representing the official policies, either expressed or implied, of the Army Research Office, Army Research Laboratory (ARL) or the U.S. Government. The U.S. Government is authorized to reproduce and distribute reprints for Government purposes notwithstanding any copyright notation herein.}